# Intelligent Mobility System with Integrated Motion Planning and Control Utilizing Infrastructure Sensor Nodes

Yufeng Yang[1], Minghao Ning[1], Shucheng Huang[1], Ehsan Hashemi[2], Amir Khajepour[1]

*Abstract*—This paper introduces a framework for an indoor autonomous mobility system that can perform patient transfers and materials handling. Unlike traditional systems that rely on onboard perception sensors, the proposed approach leverages a global perception and localization (PL) through Infrastructure Sensor Nodes (ISNs) and cloud computing technology. Using the global PL, an integrated Model Predictive Control (MPC)-based local planning and tracking controller augmented with Artificial Potential Field (APF) is developed, enabling reliable and efficient motion planning and obstacle avoidance ability while tracking predefined reference motions. Simulation results demonstrate the effectiveness of the proposed MPC controller in smoothly navigating around both static and dynamic obstacles. The proposed system has the potential to extend to intelligent connected autonomous vehicles, such as electric or cargo transport vehicles with four-wheel independent drive/steering (4WID-4WIS) configurations.

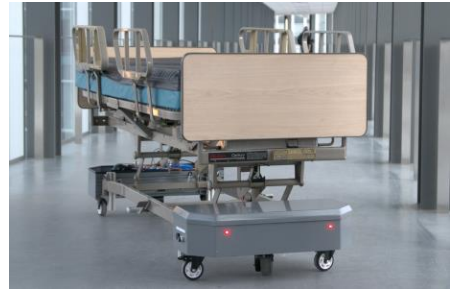

Fig. 1: Prototype of the AMR [11]

## I. INTRODUCTION

The impact of the COVID-19 pandemic on healthcare providers has been profound. According to Statistics Canada, 86.5% of healthcare providers were stressed and burned out at work during the pandemic [1]. Additionally, the World Health Organization (WHO) predicted that by 2030, there will be an 18 million nursing shortage globally [2]. This demographic shift, coupled with the potential for serious healthcare provider shortages, underlines the need for innovative solutions for enhancing health- and long-term care transportation systems that can perform some basic logistic duties.

Autonomous mobile robots (AMRs) can travel and navigate in an unknown or partially unknown environment without human intervention [3]. Most of the AMRs utilizes conventional on-board sensors, which have limited field of view (FOV) and can cause potential collisions [4]. To tackle such issues, authors in [5] and [6] proposed algorithms to improve and evaluate the perception systems' safety. However, the occlusion issues still pose a challenge.

On the other hand, local planning and tracking are typically treated as separate algorithms in conventional approaches [7] [8] [9]. Integrated path planning and tracking algorithms combine these processes into a single algorithm. One of the methods is through the combination of the model predictive control (MPC) framework with the artificial potential field (APF). Previous work has shown the effectiveness of these approaches. In [10], the authors developed a point-wise potential function based on the distance between the autonomous vehicle and the point mass obstacles as the penalty function. Since the potential function is non-linear, nonlinear MPC (NMPC) was selected as the controller. The simulation results show that the algorithm was able to avoid static point mass obstacles. Rasekhipour et al. [11] proposed a motion planner algorithm that uses MPC combined with different shapes of APF to distinguish crossable and non-crossable obstacles and lane boundaries. Bhatt et al. [12] developed a novel rule-based trajectory prediction algorithm using the MPC and APF. The results showed promising accuracy compared with learning-based methods.

Lastly, common logistic equipment found in indoor settings typically features full steering caster wheels. This is because they often need to operate through narrow spaces, and the capability to perform omnidirectional maneuvers proved to be beneficial. Therefore, the AMRs for the proposed system are equipped with two-wheel independent drive/steering mechanism to achieve omnidirectional ability. A prototype of the robot, based on retrofitting an existing medical bed, is depicted in Fig. 1 in the Mechatronic Vehicle Systems Lab [13] at the University of Waterloo. Kinematic models were primarily used for such robots, often employing reference error methods for a linear approximation [14] for the linear MPC formulation. However, this approximation approach may lead to large errors when the robot deviates from the trajectory, especially, when dodging obstacles for an integrated planning and tracking algorithm. Furthermore, Wang et al. [15] developed trajectory and velocity planning for such a robot; however, they did not consider the longitudinal speed difference between the front and rear wheels in the body coordinate frame, which may cause tire slippage during the operation.

*This research was supported by the Natural Sciences and Engineering Research Council of Canada (NSERC), MITACS, Rogers Communications Inc. and Able Invitations.

Yufeng Yang, Minghao Ning, Shucheng Huang and Amir Khajepour are with the Department of Mechanical and Mechatronics Engineering, University of Waterloo, Waterloo, ON N2L 3G1, Canada (e-mail {f248yang, minghao.ning, s95huang, a.khajepour}@uwaterloo.ca)

Ehsan Hashemi is with the Department of Mechanical Engineering, University of Alberta, Edmonton, AB T6G1H9, Canada (e-mail: ehashemi@ualberta.ca).

To tackle these challenges, this paper introduces an innovative solution that leverages a central cloud perception and control for addressing transportation needs within healthcare facilities to reduce cost and improve safety. With this approach, primary perception sensors called ISNs will be installed in the infrastructure (e.g., on the ceiling). Compared to the perception information generated using the onboard sensors, a global PL will generate a bird's-eye view of the entire operational area without any occlusions. Moreover, the global PL and control commands are determined in the cloud and transmitted to the robot via 5G or Wi-Fi networks. This enables the adoption of advanced algorithms deployed in cloud computing.

Furthermore, an integrated local planning and tracking control framework is proposed employing the linear MPC and APF for a single robot. The APF is formulated based on the closed distance between the robot and obstacles at every step within the prediction horizon. This approach enhances the accuracy of the obstacle's APF formulation, thus, reducing the likelihood of collisions. Additionally, kinematic equations are linearized utilizing the current state information and previously generated control inputs, ensuring the approximation accuracy of the kinematic equations for the proposed integrated planning and tracking controller. Lastly, a nonlinear constraint is incorporated into the MPC framework, accounting for potential slippage resulting from differences in wheel speeds.

The **contributions** of this paper can be summarized as follows:

- An indoor autonomous transportation system using ISNs and cloud computing technology is proposed to enhance the safety and performance efficiency of the robot's operation in a dynamic environment. The proposed system could be applied to connected vehicles.

- An integrated local motion planning and tracking control framework is developed using the linear MPC and APF and implemented with an augmented kinematic model formulations and customized constraints. The proposed framework could be applied to similar four-wheel independent drive/steering (4WID-4WIS) vehicles.

The paper is structured as follows: Section II outlines the proposed framework including ISNs and cloud computation. Section III discusses the design of the motion planning and tracking algorithms for a single robot. Section IV presents and discusses the simulation results obtained from the proposed integrated controller. Finally, Section V concludes this paper and points out future directions.

## II. PROPOSED SYSTEM OVERVIEW

As illustrated in Fig. 2, each ISN will first process the raw data collected from each sensor node locally, including the perception data for the obstacles and the localization data for the robots. This data will then be sent to the cloud for a global PL calculation; this information will be utilized in decision-making, and planning and control algorithms. Additionally, the wheel speed and steering angle will be measured using the pre-installed encoders in the motors. These signals will be sent back to the cloud via 5G or Wi-Fi, serving as supplementary information for the planning and control module. Lastly, these generated control commands will be transmitted to each robot via the 5G or Wi-Fi networks.

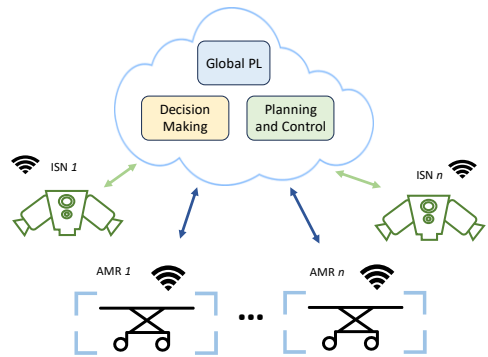

Fig. 2: Architecture of the Proposed System

To ensure the proposed robot can execute omnidirectional maneuvers, the robot's hardware design incorporates two driving modules: one attached to the front and another to the rear. As shown in Fig. 1, each module is equipped with a centrally located motorized wheel, flanked by two caster wheels on either side for balance purposes. These motorized wheels can independently generate driving and steering motion, facilitating omnidirectional movement. These two driving modules can be fitted onto any device that requires mobility.

## III. INTEGRATED MOTION PLANNING AND TRACKING

### A. Overview

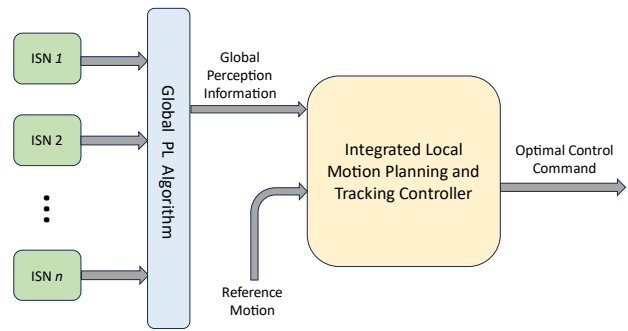

Fig. 3: Overview of the Proposed Controller

The overview of the proposed motion planning and tracking method for an individual robot is illustrated in Fig. 3, and comprises primary components: ISNs, a global PL algorithm, and an integrated local motion planning and tracking controller. Reference motion is obtained from the global planning and decision-making module, assumed known for this paper. The integrated local planning and tracking controller's task is to generate optimal control inputs enabling the robot to avoid collisions while minimizing deviation from the reference motion.

This paper utilizes linear MPC to enhance real-time efficiency. The computational burden is allocated to constructing the APF for the MPC at each step within the prediction horizon. This enables the MPC to generate the appropriate control commands for ensuring safe and smooth motion when dodging obstacles.

## B. Motion Model Description

By considering the assumptions of the rigid body and non-slippage of the tire, a three-degree-of-freedom bicycle model [15] is used to describe the motion of the proposed robot. The variables $\dot{X}$, $\dot{Y}$, and $\dot{\theta}$ represent the longitudinal velocity, lateral velocity, and rotational speed, at the robot's C.G. location in the global coordinate frame, respectively. $v_c$ and $\beta$ denote the velocity at the robot's C.G. location, and the side slip angle in the body coordinate frame. Additionally, $l_f$ and $l_r$ represent the distances from the front wheel and the rear wheel to the C.G., respectively. $v_f$, $v_r$, $\delta_f$, $\delta_r$, $a_f$, and $a_r$ denote the linear velocity, steering angle, steering angle, and linear acceleration of the front and rear wheels, respectively.

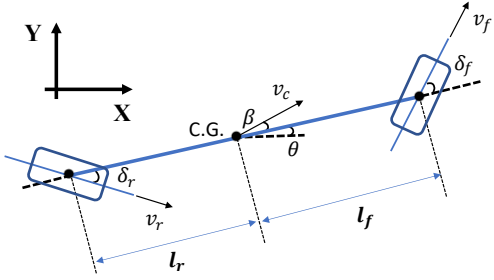

Fig. 4: Kinematic Model of AMR [15]

$$\dot{X} = v_c \cos(\theta + \beta) \quad (1)$$

$$\dot{Y} = v_c \sin(\theta + \beta) \quad (2)$$

$$\dot{\theta} = \frac{v_c \cos(\beta)[\tan(\delta_f) - \tan(\delta_r)]}{l_f + l_r} \quad (3)$$

$$\dot{v}_f = a_f \quad (4)$$

$$\dot{v}_r = a_r \quad (5)$$

$$v_c = \frac{v_f \cos(\delta_f) + v_r \cos(\delta_r)}{2\cos(\beta)} \quad (6)$$

$$\beta = \tan^{-1}\left(\frac{l_r \tan(\delta_f) + l_f \tan(\delta_r)}{l_f + l_r}\right) \quad (7)$$

In the current approach, the system state variables, $\zeta$, system outputs, $\eta$ and control inputs, $u$ are defined as follows. The reason to use linear acceleration of the wheel instead of linear velocity as the control input is because sudden changes are permissible with acceleration, as it is linked directly to the torque of the wheels.

$$\zeta = \eta = [X(t) \quad Y(t) \quad \theta(t) \quad v_f(t) \quad v_r(t)]^T \quad (8)$$

$$u = [a_f(t) \quad a_r(t) \quad \delta_f(t) \quad \delta_r(t)]^T \quad (9)$$

The above non-linear kinematic equations are linearized using the operating points: $\zeta_0(k)$, and $u_0(k)$, such as current state and previous generated control inputs. The linearized kinematic equation in the discrete time domain can be expressed as Eq. (10) to (13).

$$\begin{cases} \zeta(k+1) = A_k \zeta(k) + B_k u(k) + d_k(k) \\ \eta(k+1) = C_k \zeta(k+1) \end{cases} \quad (10)$$

$$C_k = I_5 \quad (11)$$

$$d_k(k) = \hat{\zeta}(k+1) - A_k \zeta_0(k) - B_k u_0(k) \quad (12)$$

$$\hat{\zeta}(k+1) = \zeta(k) + T_s \times f(\zeta_0(k), u_0(k)) \quad (13)$$

Since $u(k) = u(k-1) + \Delta u(k)$, the previous linearized equations can be converted to Eq. (14) that uses $\Delta u$ as the control inputs, where $\tilde{\zeta}(k) = \begin{bmatrix} \zeta(k) \\ u(k-1) \end{bmatrix}$, $\widetilde{A_k} = \begin{bmatrix} A_k & B_k \\ 0_{N_u \times N_\zeta} & I_{N_u} \end{bmatrix}$, $\widetilde{B_k} = \begin{bmatrix} B_k \\ I_{N_u} \end{bmatrix}$, $\widetilde{d_k} = \begin{bmatrix} d_k \\ 0_{N_u \times 1} \end{bmatrix}$, and $\widetilde{C_k} = [C_k \quad 0_{N_\zeta \times N_u}]$. $N_\zeta$ represents the number of system states, and $N_u$ represents the number of control inputs. The superscript of zeros and the identity matrix represents its corresponding size.

$$\begin{cases} \tilde{\zeta}(k+1) = \widetilde{A_k} \tilde{\zeta}(k) + \widetilde{B_k} \Delta u(k) + \widetilde{d_k}(k) \\ \eta(k+1) = \widetilde{C_k} \tilde{\zeta}(k+1) \end{cases} \quad (14)$$

## C. Artificial Potential Field for Motion Planning

All obstacles are assumed to be non-crossable and have a convex shape in this paper. Their potential field function is defined as an inverse function that considers both the robot's position and obstacle's position in the ground-fixed frame, which is similar to [11]. To accommodate the shape of the robot and the obstacles, a minimum distance between them is used to determine the potential field value. Additionally, the formulation of the boundaries follows the same manner but with a different priority than the obstacles by assigning different scaling constants in the APF formulation. The expression of the proposed APF function is provided as follows.

$$APF = \frac{a}{[(xx_i - xx)^2 + (yy_i - yy)^2]^b} \quad (15)$$

$$xx = X + p \quad (16)$$

$$yy = Y + q \quad (17)$$

In Eq. (15), $a$ and $b$ are the scaling constants of the APF. $(xx_i, yy_i)$ denote the location that is closest to the robot on the $i^{th}$ obstacle. Meanwhile, $(xx, yy)$ is the location of the closest distance to the $i^{th}$ obstacle on the robot; their expressions are shown in Eq. (16) and (17), where $X$ and $Y$ denote the C.G. location of the robot. The variables $p$ and $q$ represent the offset length from $(X, Y)$ to $(xx, yy)$, which can be further explained graphically in Fig. 5. In this figure, the orange box represents the robot, while the blue box represents the $i^{th}$ obstacle, respectively. $\theta$ represent the heading angle; $w$ and $l$ represents the width and height, respectively; The red and blue cross denote the locations where the robot and $i^{th}$ obstacle have the minimum distance.

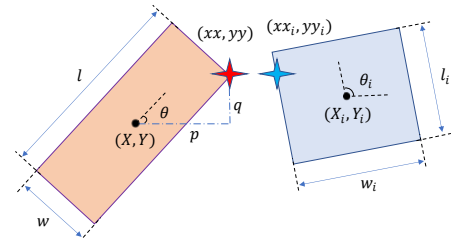

Fig. 5: Graphical Representation of APF Parameters

Since Eq. (15) is a non-linear and non-convex function, it cannot be directly included to the objective function of the linear MPC. Therefore, the defined APF approximated into a quadratic form. The corresponding gradient vector and hessian matrix can then be evaluated using the values of $(X, Y)$, and $(xx_i, yy_i)$. To ensure the convexity of the optimization problem of the MPC, the hessian matrix is converted to its nearest positive semi-definite in terms of the Frobenius norm [11] by replacing negative values in the diagonal matrix with zero from its eigendecomposition. Since the approximation of APF needs to be obtained at every time step within the prediction horizon, predicting the position of the robot and obstacles is important. In this paper, the future position of the robot is predicted using the constant previous control inputs, and the future position of the obstacles is predicted using the constant velocity and turning rate (CVTR) method.

*C. Wheel Speed Difference Constraint*

In conventional car-like robots, wheel slippage is typically caused by factors such as high-speed cornering maneuver or a low coefficient of friction on road surfaces [16] For the proposed robot as shown in Fig. 1, however, there are two independent drive wheels with independent steering angles, and tire slippage can occur if their velocities do not match the bed's rigid body motion. To prevent this, a speed constraint is defined as follows.

$$\left| v_f(k)\cos\left(\delta_f(k)\right) - v_r(k)\cos(\delta_r(k)) \right| = 0 \quad (18)$$

Since the above equation is non-linear, it is required to be linearized around the operating point $\zeta_0(k)$ and $u_0(k)$ using the Taylor series expansion. The simplified equation is shown below, where $E_k$ is the gradient vector with respect of $u$, and $g$ is Eq. (18) evaluated at the operating point. To ensure the MPC was able to obtain a feasible solution, the proposed constraint is converted into an inequality equation by limiting the range to a small number, such as $\pm 0.1$ in this paper.

$$\left| E_k \Delta u(k) + g(\zeta_0(k), u_0(k)) \right| \leq 0.1 \quad (19)$$

Note that due to the linearization process of this nonlinear constraint, even though the linearization constraint is satisfied with the obtained optimal control inputs, the nonlinear constraint may sometimes violate the defined range. However, such violation is so small that it can be ignored.

*D. Predictive Controller Design*

The MPC's formulation for the integrated local motion planning and tracking module is presented as follows. Eq. (20) is the MPC's objective function, where the first term is the APF cost; the second term is the tracking cost; and the third term is the control effort cost. The $Q$ and $R$ are the weighting matrices. Furthermore, Eq. (21) to (26) are the MPC's constraints. Specifically, Eq. (21) to (22) are the linearized kinematic equations of the robot; Eq. (23) is the limit on the changing rate of the control input; Eq. (24) is the limit on the control inputs; Eq. (25) is the limit on the system outputs; and lastly, Eq. (26) is the wheel speed difference constraint.

$$J = \left\{ \sum_{i=0}^{N_p-1} APF(k+i+1) + \left\| \eta(k+i+1) - \eta_{Ref}(k+i+1) \right\|_Q^2 + \sum_{j=0}^{N_c-1} \|\Delta u(k+j)\|_R^2 \right\} \quad (20)$$

$$\tilde{\zeta}(k+i+1) = \widetilde{A_k}\tilde{\zeta}(k+i) + \widetilde{B_k}\Delta u(k+j) + \widetilde{d_k}(k+i) \quad (21)$$

$$\eta(k+i+1) = \widetilde{C_k}\tilde{\zeta}(k+i+1) \quad (22)$$

$$\Delta u_{min} \leq \Delta u(k+j) \leq \Delta u_{max} \quad (23)$$

$$u_{min} \leq u(k+j) \leq u_{max} \quad (24)$$

$$\eta_{min} \leq \eta(k+i) \leq \eta_{max} \quad (25)$$

$$\left| E_k \Delta u(k+j) + g(\zeta_0(k), u_0(k)) \right| \leq 0.1 \quad (26)$$

## IV. RESULTS AND DISCUSSION

In this section, the performance of the proposed integrated controller is evaluated through two simulation scenarios. By taking advantage of global perceptions, obstacles' estimated position and their velocity are assumed to be known for the MPC. The parameters of the robot and the proposed MPC controller can be found in Table I. All simulations are performed using the Python platform, and computations are carried out on a Windows-based desktop equipped with an AMD Ryzen 7 2700x 4.3 GHz CPU and 32GB of RAM.

TABLE I. ROBOT AND MPC TUNING PARAMETERS

| Symbol | Value | Symbol | Value |
| --- | --- | --- | --- |
| $l_f, l_r$ | $1.2\ m$ | $N_p$ | 20 |
| $u_{max}$ | $[1,1,\frac{\pi}{2},\frac{\pi}{2}]$ | $N_c$ | 10 |
| $u_{min}$ | $-[1,1,\frac{\pi}{2},\frac{\pi}{2}]$ | $Q$ | $diag(2,2,6,10,10)$ |
| $\Delta u_{max}$ | $[0.8,0.8,\frac{\pi}{12},\frac{\pi}{12}]$ | $R$ | $100 \cdot diag(3,3,4,4)$ |
| $\Delta u_{min}$ | $-[0.8,0.8,\frac{\pi}{12},\frac{\pi}{12}]$ | $a_{obs}$ | 3 |
| $\eta_{max}$ | $[Inf, Inf, Inf, 1.4, 1.4]$ | $b_{obs}$ | 1.8 |
| $\eta_{min}$ | $-[Inf, Inf, Inf, 0.1, 0.1]$ | $a_{boundary}$ | 0.3 |
| $T_s$ | $0.1\ sec$ | $b_{boundary}$ | 1.1 |

*A. Straight Corridor*

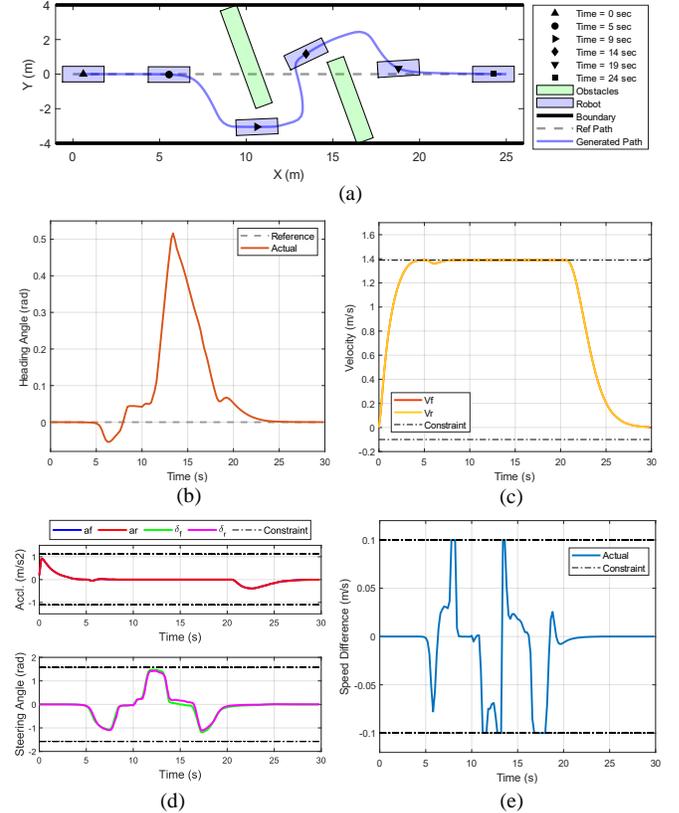

Fig. 6: Simulation Results for a Straight Corridor Scenario. (a) Motion Trajectory of robot (blue) and Obstacles (green) (b) Heading Angle of the Robot (c) Front and Rear Wheel Velocity (d) Generated Optimal Control Inputs from MPC (e) Front and Rear Wheel Speed Difference

In this scenario, the robot encounters two static obstacles that have a fixed heading angle of 60 degrees that obstruct its reference path. The reference trajectory was defined as the center line of the hallway; the reference heading angle was defined as 0 degrees; and lastly, the reference speed of two wheels was defined as a constant of 5 $km/h$.

In Fig. 6 (a), the robot is denoted in blue; the solid black markers represent different time stamps of the robot. The results demonstrate the robot was able to safely dodge obstacles while maintaining minimum derivation from its reference trajectory. Fig. 6 (b) and (c) illustrate the heading angle and wheel linear velocity of the robot, respectively. The heading angle plot demonstrates the robot closely tracked the defined reference signal when there were no obstacles around and properly adjusted its heading normal to the obstacles to reduce APF cost when traveling between the obstacles. The velocity plot shows that the robot was able to track the reference wheel velocity. A proper deceleration was applied when the robot encountered the first obstacle.

Fig. 6 (d) shows the optimal acceleration and steering angle generated by the proposed MPC, all of which fall within defined constraints. Fig. 6 (e) presents the plot for the speed difference between the front and rear wheels, which also falls within defined constraints. This underscores the accuracy of the linearization process for the nonlinear constraint.

### B. Sharp Turning in Tight Orthogonal Corridor

In this scenario, the robot encounters two obstacles obstructing its reference path. A rectangle obstacle remains stationary with a fixed heading angle of 90 degrees. Another obstacle travels from bottom to the top at a constant speed, represented by a small green square shape. The reference trajectory was again defined as the center line of the hallway; the reference heading angle was defined as the travel direction of the corridor; and lastly, the reference speed of two wheels was again defined as 5 $km/h$.

Based on the results in Fig. 7 (a), the robot successfully navigates around all obstacles. Specifically, at 7th second, the robot chooses to travel upward to reduce the cost due to the stationary obstacle. Then, at around the 20th second, the MPC predicts the future positions of both the robot and obstacles when calculating the APF, which allowed the robot to compute optimal control inputs in advance, ensuring collision avoidance while maintaining trajectory smoothness.

Fig. 7 (b) and (c) illustrate the heading angle and velocity of the robot, respectively. Upon reaching the hallway corner, the MPC generates suitable control commands for a smooth 90-degree turn. On the other hand, the velocity plot shows that the robot was able to track the reference wheel velocity when there was no obstacle around it. When encountering obstacles, the MPC was able to generate deacceleration signals that slow down the robot, reducing the APF's cost and ensuring safe maneuvers.

Fig. 7 (d) shows the optimal acceleration and steering angle generated by the proposed MPC, all of which fall within defined constraints. Fig. 7 (e) presents the plot for the speed difference between the front and rear wheels, which also falls within defined constraints.

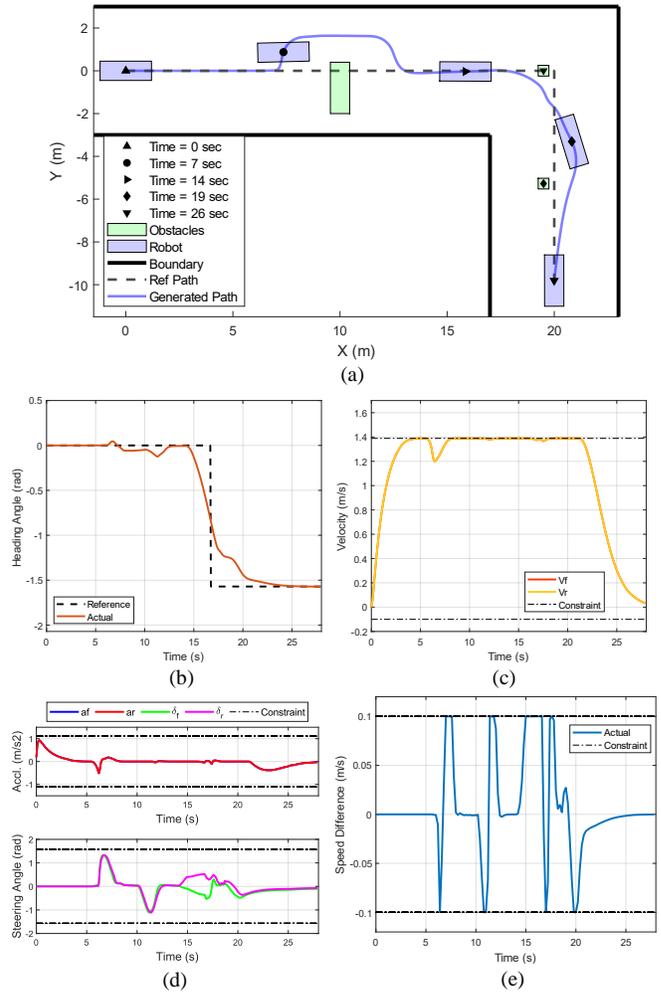

Fig. 7: Simulation Results for a 90 Degree Turning Corridor Scenario. (a) Motion Trajectory of robot (blue) and Obstacles (green) (b) Heading Angle of the Robot (c) Front and Rear Wheel Velocity (d) Generated Optimal Control Inputs from MPC (e) Front and Rear Wheel Speed Difference

### C. Discussion

In this simulation, the robot starts at the lower-left corner and is required to track the reference path while maintaining a constant heading angle of 0 degrees. Results generated by MPC without customization, i.e., prediction of the robot and obstacles, and speed difference constraints, are represented in red color, while results generated by MPC with customization are denoted in blue.

As shown in Fig. 8 (a) the MPC without the customization collided with the dynamic obstacle at the 11th second after dodging the static obstacle. This is because the MPC assumed fixed positions for both the obstacle and the robot throughout the prediction horizon at that time, leading the controller to inaccurately determine the available space to maneuver. On the other hand, by incorporating prediction, the proposed MPC anticipated the future position of dynamic obstacles within the prediction horizon, resulting the robot executed appropriate actions, such as moving upward to dodge the collisions.

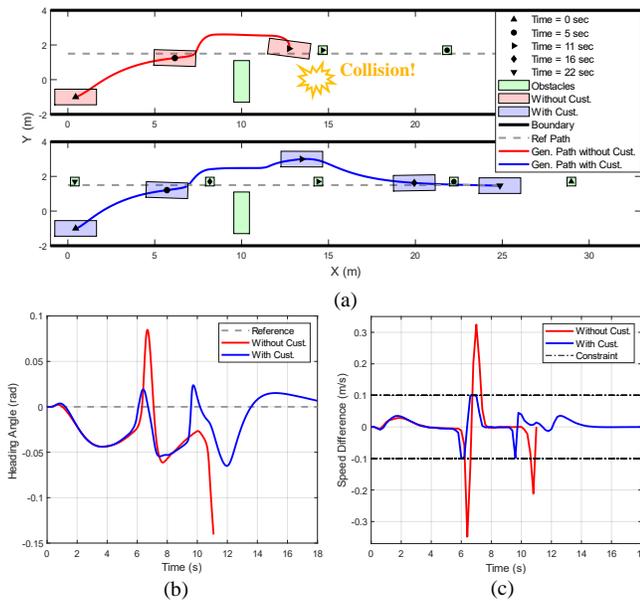

Fig. 8: Simulation Results Comparison in a Straight Corridor (a) Motion of Robot and Obstacles (b) Wheel Speed Difference (c) Heading Angle

Moreover, as illustrated in Fig. 8 (b) and (c), the MPC without customization tends to have a more aggressive change in heading angle and is more likely to have a tire slippage occurred. This is due to the controller's rapid adjustments of the steering angle, causing a violation of the non-slip assumption in the kinematic equations. In contrast, the proposed MPC smoothly adjusts the position and heading angle and satisfies the proposed speed difference constraint.

## V. CONCLUSION

This paper first presents an intelligent mobility system for indoor settings. By integrating ISNs into the infrastructure, the system acquires global PL information, thus enhancing navigation safety. Moreover, the implementation of cloud computing enables the ability to use computationally intensive algorithms for improved perception and planning strategies.

Furthermore, an integrated MPC-based local motion planning and tracking controller is developed for the proposed intelligent system to navigate the robot around obstacles. Modifications to the system state and control inputs of the robot enable linearization of the kinematic model using the current state and previous control inputs. Additionally, a nonlinear constraint is introduced and implemented into the MPC to mitigate potential speed differences between the front and rear wheels. Simulation results verify the proposed control framework, which enables the robot to avoid collisions with both stationary and dynamic obstacles while generating smooth motion. The proposed system could be potentially applied to intelligent connected autonomous vehicles, i.e., electric and transportation vehicles equipped with the 4WID-4WIS configurations.

Real-world experiments with the proposed system will be conducted to verify its effectiveness and efficiency in future works. Additionally, a learning-based prediction algorithm will be incorporated to replace the current CVTR approach to allow robot behaviour naturally especially when encountering humans.

## VI. ACKNOWLEDGEMENT

The authors would like to acknowledge the financial support of the Natural Sciences and Engineering Research Council of Canada (NSERC), and MITACS, and the financial and technical support of Rogers Communications Inc. and Able Invitations.